\documentclass[10pt,twocolumn,letterpaper]{article}

\usepackage[pagenumbers]{cvpr} % To force page numbers, e.g. for an arXiv version

\definecolor{cvprblue}{rgb}{0.21,0.49,0.74}
\usepackage[pagebackref,breaklinks,colorlinks,allcolors=cvprblue]{hyperref}
\newcommand{\myparagraph}[1]{\vspace{1mm}\noindent{\bf #1}}

\def\paperID{16} % *** Enter the Paper ID here
\def\confName{CVPR}
\def\confYear{2026}

\title{Rethinking Graph Convolution for 2D-to-3D Hand Pose Lifting}

\author{
    Chanyoung Kim\textsuperscript{\rm 1} 
    \quad
    Donghyun Kim\textsuperscript{\rm 1, 2} 
    \quad 
    Dong-Hyun Sim\textsuperscript{\rm 3} 
    \quad 
    Seong Jae Hwang\textsuperscript{\rm 2}
    \quad
    Youngjoong Kwon\textsuperscript{\rm 1} 
    \\
    \textsuperscript{\rm 1}Emory University
    \quad
    \textsuperscript{\rm 2}Yonsei University
    \quad
    \textsuperscript{\rm 3}WHATs Lab
    \\ {\tt\small \{firstname.lastname\}@emory.edu}
}

\begin{document}

\maketitle

% ============================================================
\begin{abstract}
% Graph convolutional networks (GCNs) are widely adopted for 3D hand pose estimation, encoding the hand skeleton as a fixed adjacency graph.
% We question whether GCN is the best way to leverage skeleton topology for 2D-to-3D lifting.
% In this paper, we discuss through controlled, parameter-matched ablations on the FPHA benchmark, we show that standard multi-head self-attention \textbf{outperforms GCN by 18\%} in MPJPE even when the GCN uses multi-hop adjacency and matched parameters.
% A skeleton-constrained graph attention network (GAT) recovers most of the gap (10.99\,mm), confirming that \emph{input-dependent aggregation} is the primary advantage; removing the skeleton edge constraint (full attention, 10.09\,mm) yields a further 8\% gain.
% Skeleton topology is most effective when provided as positional encoding for attention, not as fixed adjacency for GCN.
Graph convolutional networks (GCNs) are widely used for 3D hand pose estimation, where the hand skeleton is encoded as a fixed adjacency graph. We revisit whether this is the most effective way to incorporate hand topology in 2D-to-3D lifting. In this paper, we perform controlled, parameter-matched ablations on the FPHA benchmark and show that standard multi-head self-attention consistently outperforms GCN baselines. Even when the GCN is strengthened with multi-hop adjacency and matched parameter count, self-attention reduces MPJPE from 12.36 mm to 10.09 mm. A skeleton-constrained graph attention network recovers most of this gap, indicating that input-dependent aggregation is a major source of improvement, while fully connected attention yields additional gains. We further show that hand topology is most effective when introduced as a soft structural prior through graph-distance positional encoding, rather than as a hard adjacency constraint. These results suggest that, for hand pose lifting, adaptive spatial attention is a more effective inductive bias than fixed graph convolution.
\end{abstract}

% ============================================================
\section{Introduction}

Estimating 3D hand pose from egocentric views is a core capability for AR/VR interaction~\cite{fpha,assemblyhands}.
The 2D-to-3D lifting paradigm first detects 2D keypoints in the image and then regresses their 3D coordinates. It has proven effective for both body~\cite{poseformer,motionbert} and hand pose estimation~\cite{gtignet}, as it decouples appearance modeling from geometric reasoning.

Within this paradigm, graph convolutional networks (GCNs)~\cite{gcn} have become the standard tool for modeling hand structure.
Methods like HandGCNFormer~\cite{handgcnformer} and GTIGNet~\cite{gtignet} construct adjacency matrices from the hand skeleton and apply graph convolutions to propagate information between connected joints.
While effective, GCNs use a fixed adjacency structure, so every input is processed through the same connectivity pattern.
Self-attention, by contrast, computes pairwise affinities that adapt to each input: during a pinch, the thumb tip can attend to the index tip; when resolving depth ambiguity, the wrist can aggregate context from all fingertips.

\begin{figure}[t]
\centering
\includegraphics[width=\linewidth]{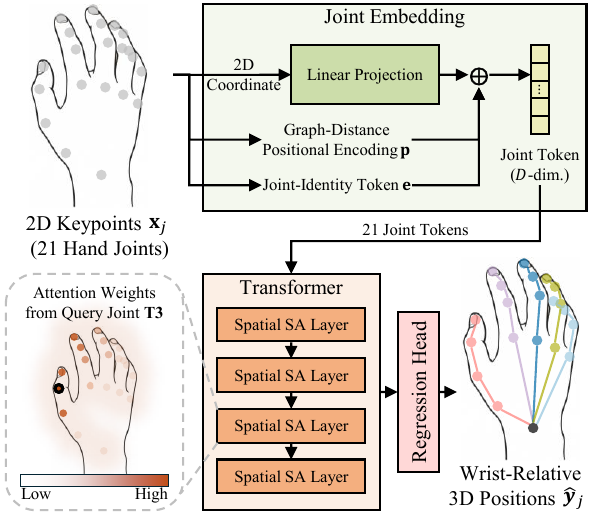}
\vspace{-10pt}
\caption{Architecture overview. 2D keypoints are embedded with learnable joint-identity tokens and graph-distance positional encoding, processed by $L{=}4$ spatial self-attention layers, and decoded to 3D coordinates via a per-joint MLP.}
\label{fig:architecture}
\vspace{-10pt}
\end{figure}

While self-attention has already proven effective for body pose lifting~\cite{poseformer,motionbert}, the hand domain presents distinct challenges that merit independent investigation.
Hands have 21 densely-packed joints in a compact spatial region, with a tree-structured kinematic chain that differs fundamentally from the body skeleton's parallel limbs.
Self-occlusion is far more severe, as fingertips routinely disappear behind the palm or other fingers during grasping, and the egocentric viewpoint introduces perspective distortion that is absent in typical body pose setups.
These differences make it unclear whether the GCN-vs-attention tradeoff observed for body pose transfers to hands.

We study this question using a deliberately minimal architecture with a linear joint embedding, graph-distance positional encoding, four Transformer blocks over 21 joint tokens, and a regression head. The model excludes any CNN backbone, graph convolution, or temporal modeling. To isolate the lifting component from 2D detection confounds, we use oracle 2D keypoints in all experiments, following standard practice in lifting research~\cite{poseformer,motionbert}.
Under this controlled setting, our model achieves 10.09$\pm$0.18\,mm MPJPE and 80.0$\pm$0.3\% AUC on FPHA over 5 random seeds.
Our carefully controlled ablations demonstrate that (1) attention outperforms GCN even when parameter count and receptive field are matched (10.09 vs.\ 12.36\,mm), and (2) skeleton topology is most effective as positional encoding for attention, not as fixed adjacency for GCN.

% ============================================================

\begin{table}[t]
\centering
\caption{Spatial block ablation on FPHA dataset. All variants share the same training setup; attention achieves the best performance among all spatial modeling choices.}
\label{tab:gcn_ablation}
\vspace{-5pt}
\small
\resizebox{\columnwidth}{!}{
\begin{tabular}{llccc}
\toprule
 & Spatial Block & MPJPE ($\downarrow$) & AUC ($\uparrow$) & Params \\
\midrule
(a) & GCN (1-hop, $D{=}256$) & 13.82{\scriptsize$\pm$0.06} & 73.0{\scriptsize$\pm$0.1} & 2.4M \\
(b) & GCN (1-hop, $D{=}296$) & 13.69{\scriptsize$\pm$0.12} & 73.3{\scriptsize$\pm$0.2} & 3.2M \\
(c) & GCN (multi-hop, $D{=}296$) & 12.36{\scriptsize$\pm$0.21} & 75.7{\scriptsize$\pm$0.4} & 3.2M \\
(d) & GAT (skeleton) & 10.99{\scriptsize$\pm$0.51} & 78.5{\scriptsize$\pm$1.0} & 3.2M \\
\midrule
(e) & \textbf{Attention} ($D{=}256$) & \textbf{10.09}{\scriptsize$\pm$0.18} & \textbf{80.0}{\scriptsize$\pm$0.3} & 3.2M \\
\bottomrule
\end{tabular}
}
\vspace{-10pt}
\end{table}

% ============================================================
\section{Method}

\myparagraph{Problem Formulation}
Given $J{=}21$ hand joint annotations in 3D world coordinates $\mathbf{y}_j^* \in \mathbb{R}^3$, we first convert to wrist-relative coordinates by subtracting the wrist position: $\mathbf{y}_j^* \leftarrow \mathbf{y}_j^* - \mathbf{y}_0^*$.
This removes global translation and focuses the model on hand articulation.
2D keypoints $\mathbf{x}_j \in \mathbb{R}^2$ are obtained by projecting the 3D annotations onto the depth camera plane using the provided intrinsics and normalizing to $[-1,1]$.
The task is to predict wrist-relative 3D positions $\hat{\mathbf{y}}_j \in \mathbb{R}^3$ from the 2D input.

\subsection{Architecture}

The model (see Fig.~\ref{fig:architecture}) consists of three stages: joint embedding, spatial self-attention, and a regression head.

\myparagraph{Joint Embedding.}
Each 2D coordinate is linearly projected and combined with a learnable joint-identity token as
\begin{equation}
    \mathbf{h}_{j}^{(0)} = \text{LN}\!\left(W_e \mathbf{x}_{j} + \mathbf{e}_j + \mathbf{p}_j\right),
\end{equation}
where $W_e \in \mathbb{R}^{D \times 2}$ is a linear projection, $\mathbf{e}_j \in \mathbb{R}^D$ is a learnable joint-identity token, $\mathbf{p}_j \in \mathbb{R}^D$ is a graph-distance positional encoding, and LN denotes LayerNorm.
The joint-identity tokens identify \emph{which} joint is being processed.
We additionally add a graph-distance positional encoding~\cite{graphormer} that modulates embeddings by shortest-path distance from the wrist along the hand skeleton.

\myparagraph{Spatial Self-Attention.}
The $J$ joint tokens pass through $L$ identical transformer layers.
Each layer applies pre-norm multi-head self-attention~\cite{attention} followed by a feed-forward network (FFN) as
\begin{align}
    \mathbf{z}^{(\ell)} &= \mathbf{h}^{(\ell)} + \text{MHA}\!\left(\text{LN}(\mathbf{h}^{(\ell)})\right), \\
    \mathbf{h}^{(\ell+1)} &= \mathbf{z}^{(\ell)} + \text{FFN}\!\left(\text{LN}(\mathbf{z}^{(\ell)})\right).
\end{align}
The multi-head attention computes queries, keys, and values from the joint tokens: $Q, K, V = W_Q \mathbf{h}, W_K \mathbf{h}, W_V \mathbf{h}$, with attention weights $\text{softmax}(QK^\top / \sqrt{d_k})$.
Each joint attends to all 21 joints, producing input-dependent affinity weights that adapt to each hand configuration.
The FFN consists of two linear layers with GELU activation and dropout: $\text{FFN}(\mathbf{x}) = W_2 \, \sigma(W_1 \mathbf{x})$, with hidden dimension $4D$.

\myparagraph{Regression Head.}
After $L$ attention layers, a per-joint MLP maps each token to a 3D coordinate:
\begin{equation}
    \hat{\mathbf{y}}_j = W_o \, \sigma\!\left(W_r \, \text{LN}(\mathbf{h}_j^{(L)})\right)
\end{equation}
where $W_r \in \mathbb{R}^{D/2 \times D}$, $W_o \in \mathbb{R}^{3 \times D/2}$, and $\sigma$ is GELU.
The full model uses $D{=}256$, $L{=}4$, 8 attention heads.

\begin{table}[t]
\centering
\caption{Positional encoding ablation. Same attention model and recipe as Table~\ref{tab:gcn_ablation}. Graph-distance PE achieves the best result.}
\label{tab:pe_ablation}
\vspace{-5pt}
\small
\resizebox{\columnwidth}{!}{
\begin{tabular}{clccc}
\toprule
 & PE Type & Topology & MPJPE ($\downarrow$) & AUC ($\uparrow$) \\
\midrule
(a) & \textbf{Graph-distance} & \checkmark & \textbf{10.09}{\scriptsize$\pm$0.18} & \textbf{80.0}{\scriptsize$\pm$0.3} \\
(b) & None (token only) & \texttimes & 10.38{\scriptsize$\pm$0.17} & 79.5{\scriptsize$\pm$0.3} \\
(c) & Learnable & \texttimes & 10.50{\scriptsize$\pm$0.17} & 79.2{\scriptsize$\pm$0.3} \\
\bottomrule
\end{tabular}
}
\vspace{-10pt}
\end{table}

\subsection{Training}

We minimize an L1 loss on wrist-relative 3D joint positions as $\mathcal{L} = \frac{1}{J}\sum_{j=1}^{J} \|\hat{\mathbf{y}}_j - \mathbf{y}_j^*\|_1$.
We use AdamW~\cite{adamw} (lr$=$1e-3, weight decay 0.01) with cosine annealing over 100 epochs, and gradient clipping at norm 1.0.
Data augmentation includes horizontal flips (with x-coordinate sign inversion to maintain consistency) and random scaling ($0.9$--$1.1\times$).
Training converges in ${\sim}$40 epochs on a single GPU.

% ============================================================
\section{Experiments}

\begin{figure}[t]
\centering
\includegraphics[width=\linewidth]{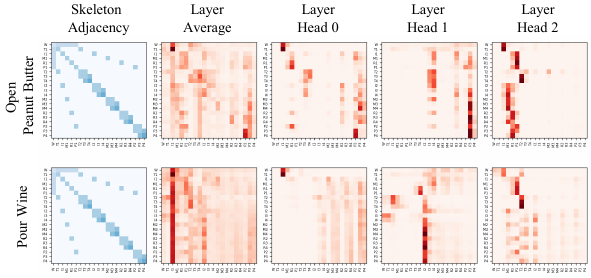}
\vspace{-15pt}
\caption{Fixed skeleton adjacency versus learned attention maps for two example actions. Unlike the sparse, input-invariant skeleton graph, attention is denser, head-specific, and pose-dependent.}
\label{fig:attn_analysis}
\vspace{-5pt}
\end{figure}
% \subsection{Setup}

\myparagraph{Dataset.}
FPHA~\cite{fpha} contains 100K+ RGB-D frames of 45 hand actions from 6 subjects with 21-joint 3D annotations.
We use the standard cross-subject split: Subjects 1, 3, 4 for training; Subjects 2, 5, 6 for testing.

\myparagraph{Evaluation and Input Protocol.}\label{sec:setup}
We report MPJPE (mm) on wrist-relative coordinates and AUC of PCK (Percentage of Correct Key-points) with thresholds 0--50\,mm.
To isolate the lifting component from 2D detection confounds, all experiments use oracle 2D keypoints, obtained by projecting GT 3D joints to 2D with known camera intrinsics, following the standard evaluation protocol in body pose lifting~\cite{poseformer,motionbert}.

\vspace{-5pt}

\subsection{Ablation: Attention vs.\ Graph Convolution}

Table~\ref{tab:gcn_ablation} compares GCN and attention-based spatial blocks under a controlled setup.
All variants share the same joint embedding, regression head, optimizer, and training schedule, so the comparison isolates the choice of spatial block.

For the GCN baselines, each block aggregates neighbor features with a symmetrically normalized adjacency matrix, $\hat{A} = D^{-1/2}(A+I)D^{-1/2}$, followed by a linear transform and ReLU:
$\mathbf{h}' = \mathrm{ReLU}(W \hat{A}\mathbf{h})$.
We begin with a standard 1-hop GCN in (a), then widen it in (b) to $D{=}296$ to match attention in parameter count.
We next enlarge the receptive field in (c) by replacing $\hat{A}$ with $\hat{A} + \hat{A}^2$, so that each layer aggregates both 1-hop and 2-hop neighbors.
To disentangle input-dependent weighting from unrestricted connectivity, we further evaluate a skeleton-constrained GAT~\cite{gat} in (d), which preserves the skeleton graph but replaces fixed graph aggregation with attention-based weighting.
Finally, full self-attention in (e) removes the edge constraint entirely and allows each joint to attend to all others.

The results reveal a clear pattern.
Relative to (a), widening the GCN in (b) yields only a marginal gain (13.82$\rightarrow$13.69\,mm), suggesting that parameter count is not the main bottleneck.
Expanding the receptive field in (c) gives a larger improvement (13.69$\rightarrow$12.36\,mm), confirming the importance of broader spatial context.
Replacing fixed graph aggregation in (c) with skeleton-constrained attention in (d) reduces error further to 10.99\,mm, showing that input-dependent weighting matters more than fixed propagation alone.
The best result is achieved by full attention in (e), indicating that unrestricted joint-to-joint connectivity provides additional benefit beyond the skeleton constraint.

\begin{figure}[t]
\centering
\includegraphics[width=\linewidth]{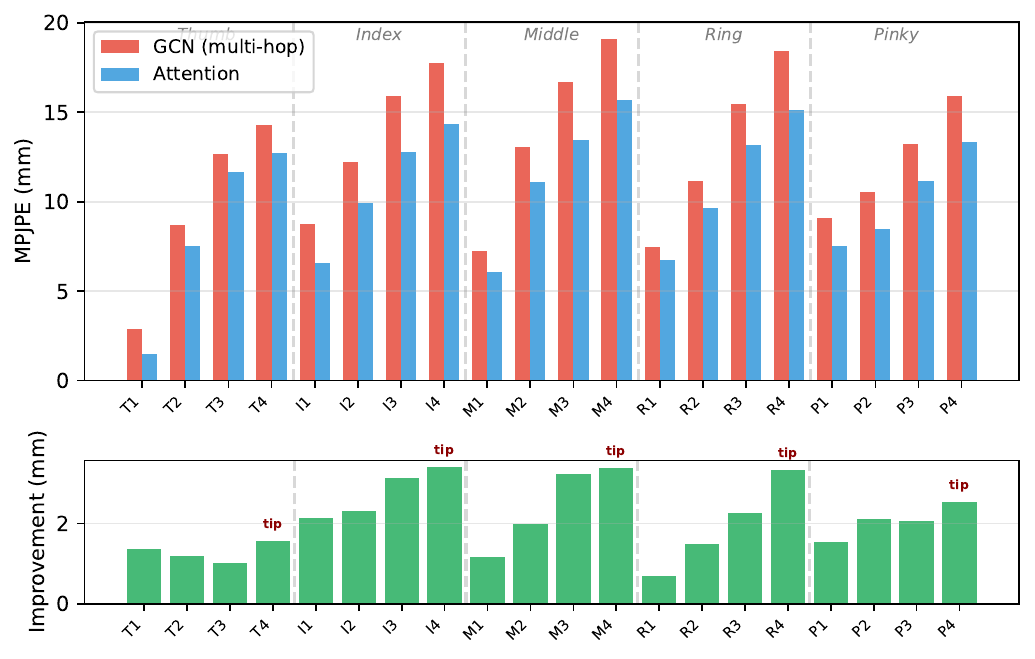}
\vspace{-20pt}
\caption{Per-joint MPJPE comparison. Attention (blue) outperforms GCN (red) on all joints, with the largest gains at fingertips (marked ``tip''), which are most prone to occlusion.}
\label{fig:perjoint}
\vspace{-5pt}
\end{figure}

\subsection{Ablation: Positional Encoding}
\label{sec:pe_ablation}

Table~\ref{tab:pe_ablation} compares three positional encoding variants for the attention model: (a) graph-distance PE, (b) no PE beyond joint-identity tokens, and (c) learnable PE without topology.

The results show that graph-distance PE in (a) performs best, achieving 10.09\,mm MPJPE and 80.0 AUC.
Removing topology, either by using no PE in (b) or learnable PE in (c), degrades performance to 10.38\,mm and 10.50\,mm, respectively.
This indicates that hand topology is useful, but more importantly, that the way it is introduced matters.

Notably, attention with graph-distance PE in (a) still substantially outperforms the parameter-matched multi-hop GCN in Table~\ref{tab:gcn_ablation} (c), even though both use the same underlying skeleton structure.
This suggests that hand topology is more effective as a soft positional prior for attention than as a fixed adjacency constraint for GCN, since attention can still adapt its interactions to the input when needed.

\begin{figure*}[t]
\centering
\includegraphics[width=\linewidth]{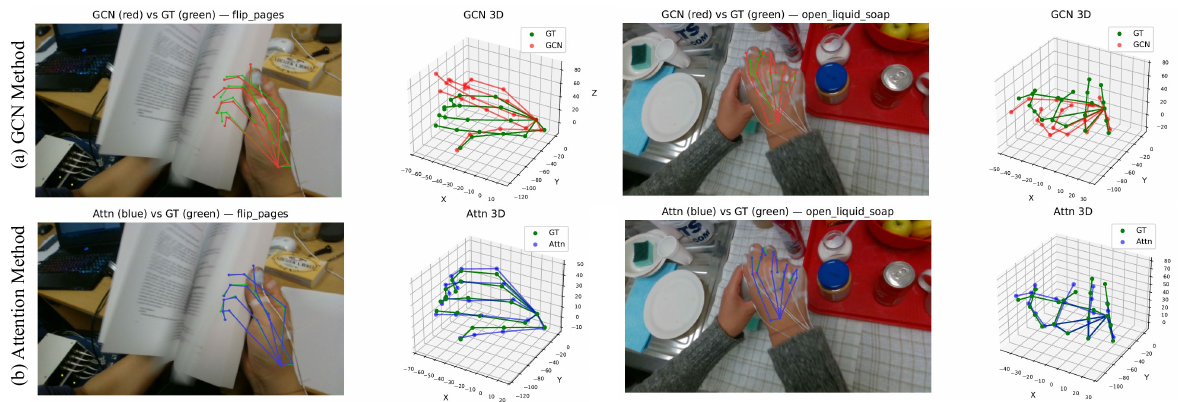}
\vspace{-20pt}
\caption{Qualitative comparison: GCN multi-hop (red) vs.\ GT (green) on the left two columns, attention (blue) vs.\ GT (green) on the right two columns. Attention consistently aligns more closely with GT, with the largest differences at fingertip positions.}
\label{fig:qualitative}
\vspace{-5pt}
\end{figure*}

\subsection{Analysis}

\myparagraph{Why Does Attention Outperform GCN?}
Table~\ref{tab:gcn_ablation} shows that GCN improves substantially when its receptive field is expanded through multi-hop adjacency, confirming that broader spatial context is crucial for hand pose lifting.
Even so, attention still outperforms the strongest GCN variant, showing that the main difference is not only how far information propagates, but how joint interactions are determined.
GCN applies the same fixed adjacency pattern to every input, whereas attention computes input-dependent affinities that adapt to the current hand configuration.

Fig.~\ref{fig:attn_analysis} highlights the difference between a fixed graph prior and adaptive attention.
The skeleton adjacency is sparse and input-invariant, imposing a single predefined interaction structure.
By contrast, the learned attention maps are denser, head-specific, and action-dependent, showing that the model dynamically reconfigures inter-joint dependencies rather than following one fixed connectivity pattern.

Quantitatively, approximately 80\% of the total attention weight falls on non-skeleton edges across all layers, with the strongest non-adjacent connections linking distal finger joints directly to the wrist.
These direct wrist connections bypass the multi-hop chain that GCN requires (e.g., 4 hops from fingertip to wrist), which explains the large fingertip gains observed in Fig.~\ref{fig:perjoint}.

\myparagraph{Per-Joint Error.}
Fig.~\ref{fig:perjoint} compares per-joint MPJPE between attention and multi-hop GCN.
Attention outperforms GCN on all 20 non-wrist joints, while the wrist error is approximately zero by definition under wrist-relative evaluation.
The improvement is largest at the fingertips (T4, I4, M4, R4, P4), where attention gains 2.84\,mm on average, compared with 1.85\,mm for non-tip joints.
These joints are both the most distal from the wrist and the most prone to occlusion in egocentric views.
This pattern suggests that attention is particularly effective where long-range context is needed, since it allows fingertip predictions to directly incorporate information from the wrist and other visible joints rather than relying on the multi-hop propagation required by GCN.

\begin{figure}[t]
\centering
\includegraphics[width=\linewidth]{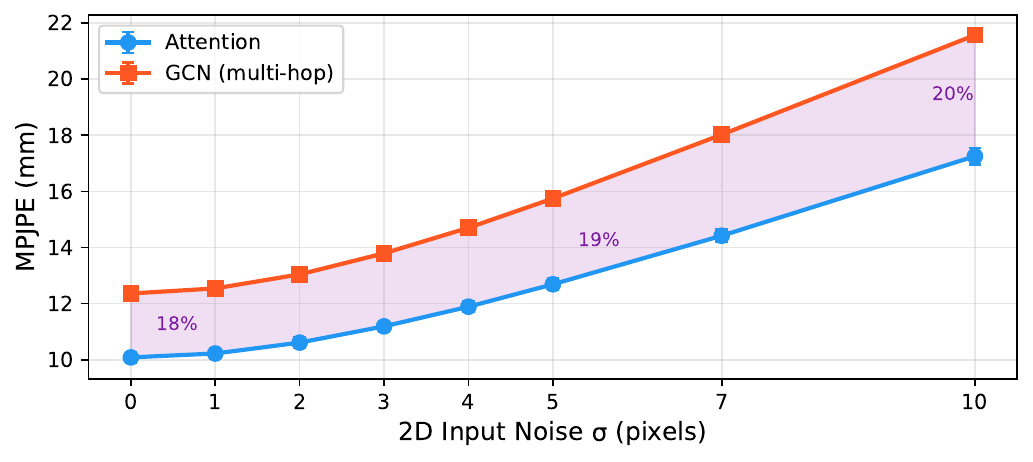}
\vspace{-20pt}
\caption{Noise robustness: oracle-trained models evaluated under increasing 2D Gaussian noise ($\sigma$\,=\,0--10\,px). Attention's advantage (shaded) grows with noise, reaching 20\% at $\sigma{=}10$\,px.}
\label{fig:noise_robustness}
\vspace{-10pt}
\end{figure}

\myparagraph{Robustness to Noisy 2D Input.}
To verify that attention's advantage is not specific to oracle input, we evaluate the same models under Gaussian noise added to the 2D keypoints at test time (Fig.~\ref{fig:noise_robustness}).
The relative gap grows from 18\% ($\sigma{=}0$) to 20\% ($\sigma{=}10$\,px), indicating that attention degrades more gracefully than GCN and that its advantage would persist with a learned 2D detector.

\myparagraph{Qualitative Comparison.}
Fig.~\ref{fig:qualitative} compares predictions from multi-hop GCN (a, red) and attention (b, blue) against ground truth (green).
Attention aligns more closely with GT, especially at fingertips, consistent with the per-joint analysis above.

\subsection{Benchmark Context}

\begin{table}[t]
\centering
\caption{FPHA benchmark context (cross-subject). $\dagger$Our model uses oracle 2D keypoints by design (Sec.~\ref{sec:setup}); results are for context only and are not directly comparable to end-to-end methods.}
\vspace{-5pt}
\label{tab:main}
\small
\resizebox{\columnwidth}{!}{
\begin{tabular}{lccc}
\toprule
Method & Input & MPJPE ($\downarrow$) & AUC ($\uparrow$) \\
\midrule
Garcia-Hern. et al.~\cite{fpha} & RGB & -- & 68.3 \\
Yang et al.~\cite{yang2022dynamic} & RGB & 11.26 & 77.5 \\
HTT~\cite{htt} & RGB & 12.13 & 76.3 \\
Roh et al.~\cite{fht} & RGB & 11.79 & 76.9 \\
GTIGNet~\cite{gtignet} & RGB & 11.15 & 77.4 \\
\midrule
Ours$^\dagger$ & Oracle 2D & 10.09{\scriptsize$\pm$0.18} & 80.0{\scriptsize$\pm$0.3} \\
\bottomrule
\end{tabular}
}
\vspace{-10pt}
\end{table}

Table~\ref{tab:main} places our lifting-only result in the context of the FPHA benchmark.
Our model uses oracle 2D input by design, so the numbers are not directly comparable to end-to-end RGB methods; the main contribution lies in the controlled ablations above.

% ============================================================
\section{Conclusion}
In this paper, we investigate whether GCN is the most effective way to model hand structure for 2D-to-3D hand pose lifting. Through controlled ablations on FPHA, we show that spatial self-attention outperforms GCN baselines even under matched parameter count and receptive field. The advantage comes from input-dependent aggregation, while hand topology is most effective as a soft structural prior through positional encoding rather than as a fixed adjacency constraint. Per-joint analysis shows the largest gains at fingertips, and a noise robustness study confirms that this advantage persists under noisy 2D input.
These findings suggest that adaptive spatial attention is a more effective spatial modeling strategy than fixed graph for hand pose lifting.

\clearpage
{
    \small
    \bibliographystyle{ieeenat_fullname}
    \bibliography{main}
}

\end{document}